# A Sentence Meaning Based Alignment Method for Parallel Text Corpora Preparation.

Krzysztof Wołk, Krzysztof Marasek

Department of Multimedia
Polish Japanese Institute of Information Technology, Koszykowa 86, 02-008 Warsaw
kwolk@pjwstk.edu.pl

**Abstract.** Text alignment is crucial to the accuracy of Machine Translation (MT) systems, some NLP tools or any other text processing tasks requiring bilingual data. This research proposes a language independent sentence alignment approach based on Polish (not position-sensitive language) to English experiments. This alignment approach was developed on the TED Talks corpus, but can be used for any text domain or language pair. The proposed approach implements various heuristics for sentence recognition. Some of them value synonyms and semantic text structure analysis as a part of additional information. Minimization of data loss was ensured. The solution is compared to other sentence alignment implementations. Also an improvement in MT system score with text processed with described tool is shown.

## 1 Introduction

Before a parallel corpus can be used for any processing, the sentences must be aligned. Sentences in the raw corpus are mostly misaligned, with translation lines whose placement does not correspond to the text lines in the source language. Moreover, some sentences may have no corresponding translation in the corpus at all. The corpus might also contain poor or indirect translations, making alignment difficult. Thus, alignment is crucial to many systems accuracy [1]. Sentence alignment must also be computationally feasible in order to be of practical use in various applications [2]. As a result, sentence alignment poses a significant challenge.

The Polish language is a particular challenge to such tools. It is a very complicated West-Slavic language with complex elements and grammatical rules. In addition, the Polish language has a large vocabulary due to many endings and prefixes changed by word declension. These characteristics have a significant effect on the requirements for the data and the data structure.

In addition, English is a position-sensitive language. The syntactic order (the order of words in a sentence) plays a very significant role, and the language has very limited inflection of words (e.g., due to the lack of declension endings). The word position in an English sentence is often the only indicator of the meaning. The sentence order follows the Subject-Verb-Object (SVO) schema, with the subject phrase preceding the predicate.

On the other hand, no specific word order is imposed in Polish, and the word order has little effect on the meaning of a sentence. The same thought can be expressed in several ways. For example, the sentence "I bought myself a new car." can be written in Polish as one of the following: "Kupiłem sobie nowy samochód";" Nowy samochód sobie kupiłem."; "Sobie kupiłem nowy samochód."; "Samochód nowy sobie kupiłem.". It must be noted that such differences exist in many language pairs and need somehow to be dealt with, what is done in this research.

This paper proposes a language independent sentence alignment method that has been applied to Polish-English parallel corpora. First, the method is described. Next, an alignment metric is discussed. A quality comparison of this method to other alignment methods is then made using data from experiments. Lastly, conclusions are drawn.

The dataset used for this research was the Translanguage English Database (TED) [12], provided by Fondazione Bruno Kessler (FBK) [17]. Vocabulary sizes of the English and Polish texts in TED are disproportionate. There are 41,684 unique English words and 88,158 unique Polish words. This also presents a challenge for SMT systems.

## 2 Literature Overview

Early attempts at automatically aligning sentences for parallel corpora were based on sentence lengths, together with vocabulary alignment [19]. Brown's method [20] was based on measuring sentence length by the number of words. Gale and Church [21] measured the number of characters in sentences. Other researchers continued exploring various methods of combining sentence length statistics with alignment of vocabularies [22, 23].

Several text aligners implementing these methods are currently available, including Bleualign, which is an open source project developed by the University of Zurich. In addition to parallel texts, Bleualign requires a translation of one of the texts. It uses the length-based Bilingual Evaluation Understudy (BLEU) similarity metric to align the texts [11].

The Hunalign tool is another open source tool, developed by the Media Research Center. Based on Gale and Church's method, it uses sentence lengths and, optionally, a dictionary to align texts in two languages strictly on a sentence level. It does not address sentence-ordering issues [10].

ABBYY Aligner is a commercial product developed by the ABBYY Group. This product reportedly uses proprietary word databases to align text portions of sentences based on their meaning. [8]

Unitex Aligner [9] is an open source project primarily developed by the University of Paris-Est Marne-la-Vallée (France). It uses the XAlign tool [18], which uses character lengths at the paragraph and sentence level for text alignment [24].

## 3 Proposed Sentence Aligner

A sentence aligner was designed to find an English translation of each Polish line in a corpus and place it in the correct place in the English file. This aligner was implemented as a Python script. We assume that each line in a text file represents one full sentence.

Our first concept is to use the Google Translator Application Programming Interface (API) for lines for which an English translation does not exist and also for comparison between the original and translated texts. The second concept is based on web crawling, using Google Translator, Bing Translator, and Babylon translator. These can work in a parallel manner to improve performance. In addition, each translator can work in many instances. Our approach can also accommodate a user-provided translation file in lieu of crowd sourcing.

Our strategy is to find a correct translation of each Polish line aided by Google Translator or another translation engine. We translate all lines of the Polish file (src.pl) with Google Translator and put each line translation in an intermediate English translation file (src.trans). This intermediate translation helps us find the correct line in the English translation file (src.en) and put it in the correct position.

In reality, the actual solution is more complex. Suppose that we choose one of the English data lines as the most similar line to the specific translated line and that its similarity rate is high enough to be accepted as the translation. This line can be more similar to the next line of src.trans, so that the similarity rate of this selected line and the next line of src.trans is higher. For example, consider the sentences and their similarity rating in Table 1.

**Table 1.** Example Similarity Ratings

| src.trans | src.en | Sim. |
|---|---|---|
| I go to school every day. | I like going to school every day. | 0.60 |
| I go to school every day. | I do not go to school every day. | 0.70 |
| I go to school every day. | We will go tomorrow. | 0.30 |
| I don't go to school every day. | I like going to school every day. | 0.55 |
| I don't go to school every day. | I do not go to school every day. | 0.95 |
| I don't go to school every day. | We will go tomorrow. | 0.30 |

In this situation, we should select "I do not go to school every day." from src.en instead of "I don't go to school every day" from src.trans, and not "I go to school every day.". So, we should consider the similarity of a selected line with the next lines of src.trans to make the best possible selection in the alignment process.

There are additional complexities that must be addressed. Comparing the src.trans lines with the src.en lines is not easy, and it becomes harder when we want to use the similarity rate to choose the correct, real-world translation.

There are many strategies to compare two sentences. We can split each sentence into its words and find the number of words in both sentences. However, this approach has some problems. For example, let us compare "It is origami." to these sentences: "The common theme what makes it origami is folding is how we create the form."; "This is origami."

With this strategy, the first sentence is more similar because it contains all 3 words. However, it is clear that the second sentence is the correct choice. We can solve this problem by dividing the number of words in both sentences by the number of total words in the sentences. However, counting stop words in the intersection of sentences sometimes causes incorrect results. So, we remove these words before comparing two sentences.

Another problem is that sometimes we find stemmed words in sentences, for example "boy" and "boys." Despite the fact that these two words should be counted as similarity of two sentences, with this strategy, these words are not counted.

The next comparison problem is the word order in sentences. In Python there are other ways for comparing strings that are better than counting intersection lengths. The Python "difflib" library for string comparison contains a function that first finds matching blocks of two strings. For example, we can use difflib to find matching blocks in the strings "abxcd" and "abcd".

Difflib's "ratio" function divides the length of matching blocks by the length of two strings, and returns a measure of the sequences' similarity as a float value in the range [0, 1]. This measure is 2.0*M / T, where T is the total number of elements in both sequences, and M is the number of matches. Note that this measure is 1.0 if the sequences are identical, and 0.0 if they have nothing in common. Using this function to compare strings instead of counting similar words helps us to solve the problem of the similarity of "boy" and "boys". It also solves the problem of considering the position of words in sentences.

Another problem in comparing lines is synonyms. For example, in these sentences: "I will call you tomorrow."; "I would call you tomorrow."

If we want to know if these sentences are the same, we should know that "will" and "would" can be used interchangeably.

We used the NLTK Python module and WordNet® to find synonyms for each word and to use these synonyms in comparing sentences. Using synonyms of each word, we created multiple sentences from each original sentence.

For example, suppose that the word "game" has the synonyms: "play", "sport", "fun", "gaming", "action", and "skittle". If we use, for example, the sentence "I do not like game.", we create the following sentences: "I do not like play."; "I do not like sport."; "I do not like fun."; "I do not like gaming."; "I do not like action."; "I do not like skittle.". We must do the same every word in a sencence.

Next, we try to find the best score by comparing all these sentences instead of just comparing the main sentence. One issue is that this type of comparison takes too much time, because we need to do many comparisons for each selection.

Difflib has other functions (in SequenceMatcher and Diff class) to compare strings that are faster than described solution, but their accuracy is worse. To overcome all these problems and obtain the best results, we consider two criteria: the speed of the comparison function and the comparison acceptance rate.

To obtain the best results, our script provides users with the ability to have multiple functions with multiple acceptance rates. Fast functions with lower quality results are tested first. If they can find results with a very high acceptance rate, we accept their selection. If the acceptance rate is not sufficient, we can use slower but higher accuracy functions. The user can configure these rates manually and test the resulting quality to get the best results. All are well described in documentation [25].

Because we used the Google Translator API and comparison functions that are not specific to any language, the program should be able to align similarly structured languages that are supported by Google Translator with English. Alignment between a language pair not included in Google Translator or WordNet would require use of a different lexical library for synonyms or not using some comparison functions.

Information about each data domain would require adapting parameters in order to provide the best alignment. In general, texts is associated with a domain, i.e. a particular subject area and mode of writing, e.g., a political science essay [13]. As discussed in [14], texts from different domains are likely to use words with different meanings. If a domain is ignored, this can lead to translations that are misleading [26].

The proposed method automatically creates text corpora. Some other aligners work in only a semi-automatic or fully manual manner. If they are unable to align or there is no translation, they leave an empty line. Clearly, this result in problems and some information are lost in process, which does not occur in our solution.

## 4 Sentence Alignment Metric

We developed a special metric to evaluate aligner quality and tuned its parameters during this research. Special metric was needed to evaluate sentences properly aligned but built from synonyms or with different phrase order. For an aligned sentence, we give 1 point. For a misaligned sentence, we give a -0.2 points penalty. For web service translations, we give 0.4 points. For translations due to disproportion between input files, we give 1 point (when one of two files included more sentences). The score is normalized to fit between 1 and 100. A higher value is better. A floor function can be used to round the score to an integer value. Point weights were determined by empirical research and can be easily adjusted if needed. The score $S$ is defined as:

$$S = floor(\frac{20(5A - M + 2T + 5|D|)}{L}) \quad (1)$$

where A is the number of aligned sentences, M is the number of misaligned sentences, T is the number of translated sentences, D is the number of lines not found in both language files (one file can contain some sentences that do not exist in the other one), and L is the total number of output lines.

Some additional scoring algorithms were also implemented. These are more suited for comparing language translation quality. We added the BLEU and native implementations of Translation Edit Rate (TER) and Character Edit Rate (CER) by using pycdec and the Rank-based Intuitive Bilingual Evaluation Measure (RIBES). BLEU and TER are well-described in the literature [4-5]. RIBES is an automatic evaluation metric for machine translation, developed in NTT Communication Science Labs [6].

The pycdec module is a Python interface to the cdec decoding and alignment algorithms [15, 16]. The BLEU metric compares phrases from a source text with reference translations of the text, using weighted averages of the resulting matches. It has been shown that BLEU performs well in comparison to reference human translations. BLEU is defined in [3] in terms of the *n*-gram precision (using *n*-grams up to length N) $p_n$ and weights $w_n$ (positive only) whose sum is one:

$$BLEU = P_B \exp(\sum_{n=0}^{N} w_n \log p_n) \qquad (2)$$

Here, $P_B$ is the brevity penalty, which is given by [3] as:

$$P_B = \begin{cases} 1, & c > r \\ e^{(1-\frac{r}{c})}, & c \leq r \end{cases} \qquad (3)$$

In the equation above, *c* is the candidate phrase translation length, and *r* is the length of the reference translation phrase, e is Euler's constant. [3].

The TER metric is intended to capture the quality of essential meaning and fluency of SMT system translations. It measures human translator edits required for a machine translation to match a reference translation. TER accounts for word substitutions, word insertions and deletions, and phrase and word order modifications [5].

We use these algorithms to generate likelihood scores for two sentences, to choose the best one in the alignment process. For this purpose, we used cdec and pycdec. cdec is a decoder, aligner, and learning framework for statistical machine translation and similar structured prediction models. It provides translation and alignment modeling based on finite-state transducers and synchronous context-free grammars, as well as implementations of several parameter-learning algorithms [15, 16].

pycdec is a Python module for the cdec decoder. It enables Python coders to use cdec's fast C++ implementation of core finite-state and context-free inference algorithms for decoding and alignment. The high-level interface allows developers to build integrated MT applications that take advantage of the rich Python ecosystem without sacrificing computational performance. The modular architecture of pycdec separates search space construction, rescoring, and inference.

cdec includes implementations of the basic evaluation metrics (BLEU, TER and CER), exposed in Python via the cdec.score module. For a given (reference, hypothesis) pair, sufficient statistics vectors (SufficientStats) can be computed. These vectors are then summed for all sentences in the corpus, and the final result is converted into a real-valued score.

Before aligning a big data file, it is important to determine the proper comparators and acceptance rates for each one. Files of 1000 – 10,000 lines result in the best performance. We recommend first evaluating each comparison method separately, and then combining the best ones in a specific scenario. For this purpose, we recommend using the binary search method in order to determine the best threshold factor value.

## 5 Comparison Experiments

Experiments were performed to compare the performance of the proposed method with several other sentence alignment implementations on the data found in [7], using the metric defined earlier. The Polish data in the Translanguage English Database (TED) lectures (approximately 15 MB) includes over 2 million words that are not tokenized.

The additional aligners, all created to develop parallel corpora, used in this experiment were: Bleualign, hunalign, ABBYY Aligner, Wordfast Aligner, and Unitex Aligner. The performance of the aligners was scored using the sentence alignment metric described in Section 3. Table 2 provides the results.

**Table 2.** Experimental Results

| Aligner | Score |
|---|---|
| Proposed Method | 98.94 |
| Bleualign | 96.89 |
| Hunalign | 97.85 |
| ABBYY Aligner | 84.00 |
| Wordfast Aligner | 81.25 |
| Unitex Aligner | 80.65 |

Clearly, the first three aligners scored well. The proposed method is fully automatic. It is important to note that Bleualign does not translate text and requires that it be done manually.

As discussed earlier, it is important not to lose lines of text in the alignment process. Table 3 shows the total lines resulting from the application of each alignment method.

**Table 3.** Experimental Results

| Aligner | Lines |
|---|---|
| Human Translation | 1005 |
| Proposed Method | 1005 |
| Bleualign | 974 |
| Hunalign | 982 |
| ABBYY Aligner | 866 |
| Wordfast Aligner | 843 |
| Unitex Aligner | 838 |

All the aligners compared, other than the proposed method, lose lines of text as compared to a reference human translation. The proposed method lost no lines.

In purpose of showing the output quality with an independent metric we decided to compare results with BLEU, NIST, METEOR and TER (the lower the better), in a comparison with human B1aligned texts. Those results are presented in Table 4.

**Table 4.** BLEU Comparison Results

| Aligner | BLEU | NIST | MET | TER | % of correctness |
|---|---|---|---|---|---|
| Human Translation | 100 | 15 | 100 | 0 | 100 |
| Proposed Method | 98,91 | 13,81 | 99,11 | 1,38 | 98 |
| Bleualign | 91,62 | 13,84 | 95,27 | 9,19 | 92 |
| Hunalign | 93,10 | 14,10 | 96,93 | 6,68 | 94 |
| ABBYY Aligner | 79,48 | 12,13 | 90,14 | 20,01 | 83 |
| Wordfast Aligner | 85,64 | 12,81 | 93,31 | 14,33 | 88 |
| Unitex Aligner | 82,20 | 12,20 | 92,72 | 16,39 | 86 |

## 6 Conclusions

In general, sentence alignment algorithms are very important for creating of parallel text corpora. Most aligners are not fully automatic, but the one proposed here is, which gives it a distinct advantage. It also allows creating a corpus when sentences exist just in a single language. The proposed approach is also language independent for ones with similar structure to PL or EN.

The results show that the proposed method performed very well in terms of the metric. It also lost no lines of text, unlike the other aligners. This is critical to the end goal of obtaining a translated text. Our alignment method also proved to provide better score when comparing with typical machine translation metrics, and would most likely improve MT systems output quality.

## 6 Acknowledgements

This work is supported by the European Community from the European Social Fund within the Interkadra project UDA-POKL-04.01.01-00-014/10-00 and Eu-Bridge 7th FR EU project (grant agreement n°287658).

## References

1. Deng Y., Kumar S. and Byrne W., "Segmentation and alignment of parallel text for statistical machine translation", *Natural Language Engineering*, 12(4), p. 1-26, 2006.
2. Braune F. and Fraser A., "Improved Unsupervised Sentence Alignment for Symmetrical and Asymmetrical Parallel Corpora", *Coling 2010: Poster Volume*, pages 81-89, August 2010.

3. Papineni, K., Rouskos, S., Ward, T., and Zhu, W.J. "BLEU: a Method for Automatic Evaluation of Machine Translation", *Proc. of 40th Annual Meeting of the Assoc. for Computational Linguistics*, Philadelphia, July 2002, pp. 311-318.
4. Snover, M., Dorr, B., Schwartz, R., Micciulla, L., and Makhoul, J., "A Study of Translation Edit Rate with Targeted Human Annotation", *Proc. of 7th Conference of the Assoc. for Machine Translation in the Americas*, Cambridge, August 2006.
5. Levenshtein, V. I. "Binary codes with correction for deletions and insertions of the symbol 1", *Problemy Peredachi Informacii*, 1965.
6. Linguistic Intelligence Research Group, NTT Communication Science Laboratories. RIBES: Rank-based Intuitive Bilingual Evaluation Score, http://www.kecl.ntt.co.jp/icl/lirg/ribes/, retrieved on August 7, 2013.
7. International Workshop on Spoken Language Translation (IWSLT), http://www.iwslt2013.org/, retrieved on August 7, 2013.
8. ABBYY Aligner, http://www.abbyy.com/aligner/, retrieved on August 7, 2013.
9. Unitex/Gramlab, http://www-igm.univ-mlv.fr/~unitex/#, retrieved on August 7, 2013.
10. hunalign – sentence aligner, http://mokk.bme.hu/resources/hunalign/, retrieved on August 8, 2013.
11. Bleualign, https://github.com/rsennrich/Bleualign, retrieved on August 8, 2013.
12. Marasek, K., "TED Polish-to-English translation system for the IWSLT 2012", Proc. of International Workshop on Spoken Language Translation (IWSLT) 2010, Hong Kong, December 2012.
13. Schmidt, A., *Statistical Machine Translation Between New Language Pairs Using Multiple Intermediaries* (Doctoral dissertation, Thesis), 2007.
14. Specia, L., Raj, D. and Turchi, M., "Machine translation evaluate versus quality estimation", *Machine Translation*, 24:39-50, 2010.
15. Chahuneau, V., Smith, N.A., and Dyer, C., pycdec: A Python Interface to cdec. *The Prague Bulletin of Mathematical Linguistics*, No. 98, 2012, pp. 51–61.
16. Dyer, C. et al.,). "cdec: A decoder, alignment, and learning framework for finite-state and context-free translation models", *Proc. of ACL 2010 System Demonstrations* (pp. 7-12). Association for Computational Linguistics, July 2010.
17. Cettolo, M., Girardi, C., & Federico, M., "Wit3: Web inventory of transcribed and translated talks", *Proc. of 16th Conference of the European Association for Machine Translation (EAMT)*, Trento, Italy (pp. 261-268), May 2012.
18. Paumier, S., Nakamura, T., & Voyatzi, S. (2009). UNITEX, a Corpus Processing System with Multi-Lingual Linguistic Resources. eLEX2009, 173.
19. Santos, A., "A survey on parallel corpora alignment", *MI-STAR 2011*, Pages 117–128.
20. Brown, P.F., Lai, J.C., and Mercer, R.L., "Aligning sentences in parallel corpora", *Proc. of 29th Annual Meeting of the ACL*, pages 169-176, Berkeley, 1991.
21. Gale, W.A., and Church, K.W., "Identifying word correspondences in parallel texts", *Proc. of DARPA Workshop on Speech and Natual Language*, pages 152-157, 1991.
22. Varga, D. et al., "Parallel corpora for medium density languages", *Proc. of the RANLP 2005*, pages 590-596, 2005.
23. Braune, F. and Fraser, A., "Improved unsupervised sentence alignment for symmetrical and asymmetrical parallel corpora", *Proc. of 23rd COLING International Conference*, pages 81-89, Beijing, China, 2010.
24. Bonhomme, P. & Romary, L., "The lingua parallel concordancing project: Managing multilingual texts for educational purpose", *Proc. of Quinzièmes Journées Internationales IA 95*, Montpellier, 1995.
25. http://korpusy.s16874487.onlinehome-server.info/
26. Thorleuchter, D. and Van den Poel, D. "Web Mining based Extraction of Problem Solution Ideas", Expert Systems with Applications, 40(10), p. 3961-3969, 2013.